\documentclass[10pt,twocolumn,letterpaper]{article}

\usepackage[pagenumbers]{cvpr} 

\usepackage{appendix}
\usepackage{pifont}
\usepackage{multirow}
\usepackage{bigstrut}
\usepackage{graphicx}
\usepackage{amsmath}
\usepackage{amssymb}
\usepackage{booktabs}
\usepackage{algpseudocode}
\usepackage{algorithmicx,algorithm}
\usepackage{times}
\usepackage{epsfig}
\usepackage{appendix}
\usepackage{mathptm}
\usepackage{multirow}
\usepackage{booktabs} 
\usepackage[table,xcdraw]{xcolor}
\usepackage{color}

\usepackage{esint}
\usepackage{newtxmath}

\usepackage[pagebackref,breaklinks,colorlinks]{hyperref}

\usepackage[capitalize]{cleveref}
\crefname{section}{Sec.}{Secs.}
\Crefname{section}{Section}{Sections}
\Crefname{table}{Table}{Tables}
\crefname{table}{Tab.}{Tabs.}

\def\cvprPaperID{793} 
\def\confName{CVPR}
\def\confYear{2023}

\begin{document}

\title{CAT: Lo\underline{C}alization and Identific\underline{A}tion Cascade Detection \underline{T}ransformer \\ for Open-World Object Detection}

\author{
    Shuailei Ma \textsuperscript{\rm 1}\thanks{Equal contribution.} ~~ 
    Yuefeng Wang\textsuperscript{\rm 1}\footnotemark[1]  ~~ 
    Ying Wei\textsuperscript{\rm 1 2}\thanks{Corresponding author.} ~ 
    Jiaqi Fan\textsuperscript{\rm 1}\\
    Thomas H. Li\textsuperscript{\rm 3} ~
    Hongli Liu\textsuperscript{\rm 4} ~
    Fanbing Lv\textsuperscript{\rm 4} \\
    \textsuperscript{\rm 1}\small{Northeast University, Shenyang, China}
    \textsuperscript{\rm 2}\small{Information Technology R\&D Innovation Center of Peking University} \\
    \textsuperscript{\rm 3}\small{School of Electronic and Computer Engineering, Peking University Shenzhen Graduate School, Shenzhen, China}\\
    \textsuperscript{\rm 4}\small{Changsha Hisense Intelligent System Research Institute Co., Ltd} \\
}

\maketitle

\begin{abstract}

Open-world object detection (OWOD), as a more general and challenging goal, requires the model trained from data on known objects to detect both known and unknown objects and incrementally learn to identify these unknown objects. The existing works which employ standard detection framework and fixed pseudo-labelling mechanism (PLM) have the following problems: $(i)$ The inclusion of detecting unknown objects substantially reduces the model's ability to detect known ones. $(ii)$ The PLM does not adequately utilize the priori knowledge of inputs. $(iii)$ The fixed selection manner of PLM cannot guarantee that the model is trained in the right direction. We observe that humans subconsciously prefer to focus on all foreground objects and then identify each one in detail, rather than localize and identify a single object simultaneously, for alleviating the confusion. This motivates us to propose a novel solution called CAT: Lo\underline{\textbf{C}}alization and Identific\underline{\textbf{A}}tion Cascade Detection \underline{\textbf{T}}ransformer which decouples the detection process via \textbf{the shared decoder} in \textbf{the cascade decoding way}. In the meanwhile, we propose the \textbf{self-adaptive pseudo-labelling mechanism} which combines the model-driven with input-driven PLM and self-adaptively generates robust pseudo-labels for unknown objects, significantly improving the ability of CAT to retrieve unknown objects. Experiments on two benchmarks, $i.e.$, MS-COCO and PASCAL VOC, show  that our model outperforms the state-of-the-art methods.
The code is publicly available at \url{https://github.com/xiaomabufei/CAT}.

\end{abstract}

\begin{figure}[htbp]
  \setlength{\belowcaptionskip}{-0.3cm}
  \centering
  \includegraphics[width = \linewidth]{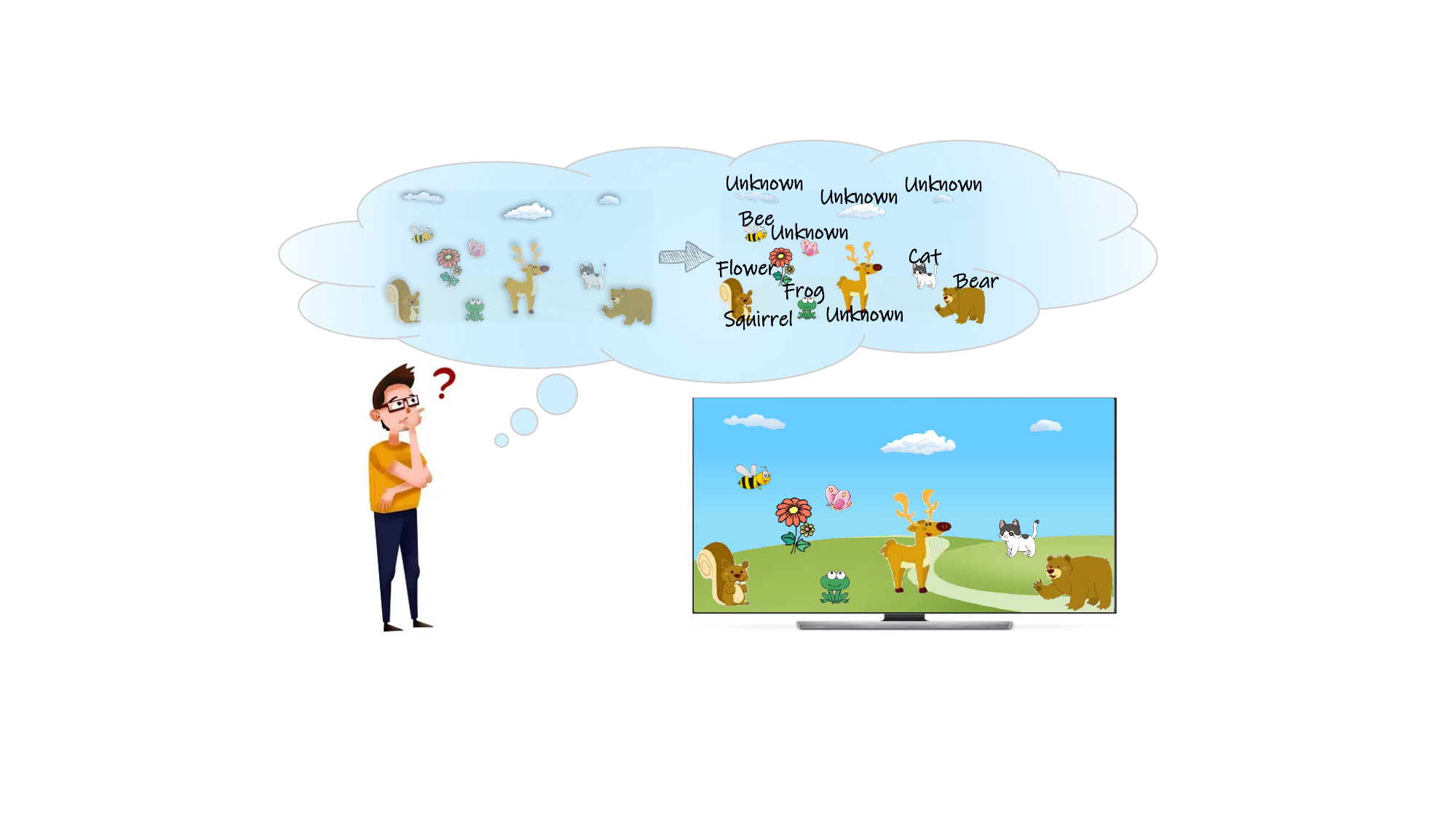}
  \caption{ When faced with new scenes in open world, humans subconsciously focus on all foreground objects and then identify them in detail in order to alleviate the confusion between the known and unknown objects and get a clear view. Motivated by this, our CAT utilizes the shared decoder to decouple the localization and identification process in the cascade decoding way, where the former decoding process is used for localization and the latter for identification.
  }
  \label{fig:Figure1}
\end{figure}
\section{Introduction}

Open-world object detection (OWOD) is a more practical detection problem in computer vision, making artificial intelligence (AI) smarter to face more difficulties in real scenes. Within the OWOD paradigm, the model’s life-span is pushed by iterative learning process. At each episode, the model trained only by known objects needs to detect known objects while simultaneously localizing unknown objects and identifying them into the unknown class. Human annotators then label a few of these tagged unknown classes of interest gradually. The model given these newly-added annotations will continue to incrementally update its knowledge  without retraining from scratch. 


Recently, Joseph \etal \cite{ORE} proposed an open-world object detector, ORE, based on the two-stage Faster R-CNN \cite{fasterrcnn} pipeline. ORE utilized an auto-labelling step to obtain pseudo-unknowns for training model to detect unknown objects and learned an energy-based binary classifier to distinguish the unknown class from known classes. However, its success largely relied on a held-out validation set which was leveraged to estimate the distribution of unknown objects in the energy-based classifier. Then, several methods \cite{UC-OWOD, OCPL, two-branch, mvit} attempted to extend ORE and achieved some success. To alleviate the problems in ORE, Gupta \etal \cite{owdetr} proposed to use the detection transformer \cite{detr,ddetr} for OWOD in a justifiable way and directly leveraged the framework of DDETR \cite{ddetr}. In addition, they proposed an attention-driven PLM which selected pseudo labels for unknown objects according to the attention scores. 

For the existing works, we find the following hindering problems. $(i)$ Owing to the inclusion of detecting unknown objects, the model's ability to detect known objects substantially drops. To alleviate the confusion between known and unknown objects, humans prefer to dismantle the process of open-world object detection rather than parallelly localize and identify open-world objects like most standard detection models. $(ii)$ To the best of our knowledge, in the existing OWOD PLM, models leverage the learning process for known objects to guide the generation of pseudo labels for unknown objects, without leveraging the prior conditions of the inputs ($texture, light\ flow, etc$). As a result, the model cannot learn knowledge beyond the data annotation. $(iii)$ The fixed selection manner of PLM cannot guarantee that the model learns to detect unknown objects in the right direction, due to the uncertain quality of the pseudo labels. The models may be worse for detecting unknown objects.

When faced with a new scene, humans prefer focusing on all foreground objects and then analysing them in detail \cite{CARRASCO20111484}, as shown in Figure.\ref{fig:Figure1}. Motivated by this and the aforementioned observations, we propose a novel Lo\underline{\textbf{C}}alization and Identific\underline{\textbf{A}}tion Cascade Detection \underline{\textbf{T}}ransformer. CAT comprises three dedicated components namely, \textbf{shared transformer decoder}, \textbf{cascade decoupled decoding manner} and \textbf{self-adaptive pseudo-labelling mechanism}. Via the cascade decoupled decoding manner, the shared transformer decoder decouples the localization and identification process. Therefore, the influence of the category information of the identification process on the localization process is reduced. In this case, the model can localize more foreground objects so that the model's ability to retrieve unknown objects is improved. Meanwhile, the independent recognition process allows the model to identify with more focus, so that the influence of unknown on detecting known objects is alleviated. In this decoding way, the former decoding process is used for localization and the latter for identification.  The self-adaptive PLM maintains the ability of CAT to explore the knowledge beyond the known objects and self-adaptively adjusts the pseudo-label generation according to the model training process. Our contributions can be summarized fourfold: 
\begin{itemize}\setlength{\itemsep}{1pt}

\item[$\bullet$]We propose a novel localization and identification cascade detection transformer (CAT), which has excellent ability to retrieve unknown objects and alleviate the influence of detecting unknown objects on the detection of known ones.

\item[$\bullet$]Inspired by the subconscious reactions when people face open scenes, we propose the cascade decoupled decoding way, which decouples the decoding procedure via the shared decoder. 

\item[$\bullet$]We introduce a novel pseudo-labelling mechanism that self-adaptively combines the model-driven and input-driven pseudo-labelling during the training process for generating robust pseudo-labels and exploring knowledge beyond known objects.

\item[$\bullet$]Our extensive experiments on two popular benchmarks demonstrate the effectiveness of the proposed CAT. CAT outperforms the state-of-the-art methods for OWOD, IOD, and open-set detection. For OWOD, CAT achieves absolute gains ranging from 9.7\% to 12.8\% in terms of unknown recall over the SOTA method.
\end{itemize}

\section{Problem Formulation}
At time $t$, let $\mathcal{K}^{t}=\{1,2, \ldots, C\}$ denote the set of known object classes and $\mathcal{U}^{t}=\{C+1, \ldots\}$ denote the unknown classes which might be encountered at the test time. The known object categories $\mathcal{K}^{t}$ are labeled in the dataset $\mathcal{D}^{t}=\left\{\mathcal{J}^{t}, \mathcal{L}^{t}\right\}$ where $\mathcal{J}^{t}$ denotes the input images and $\mathcal{L}^{t}$ denotes the corresponding labels at time $t$. The training image set consists of $M$ images $\mathcal{J}^{t}=\left\{i_{1}, i_{2}, \ldots, i_{M}\right\}$ and corresponding labels $\mathcal{L}^{t}=\left\{\ell_{1}, \ell_{2}, \ldots, \ell_{M}\right\}$. Each $\ell_{i}=\left\{\mathcal{T}_{1}, \mathcal{T}_{2}, \ldots, \mathcal{T}_{N}\right\}$ denotes a set of $N$ object instances with their class labels $c_{n} \subset \mathcal{K}^{t}$ and locations, $ x_{n}, y_{n}, w_{n}, h_{n}$ denote the bounding box center coordinates, width and height respectively. 
\begin{figure*}[htbp]
  \centering
  \includegraphics[width = 0.9\textwidth]{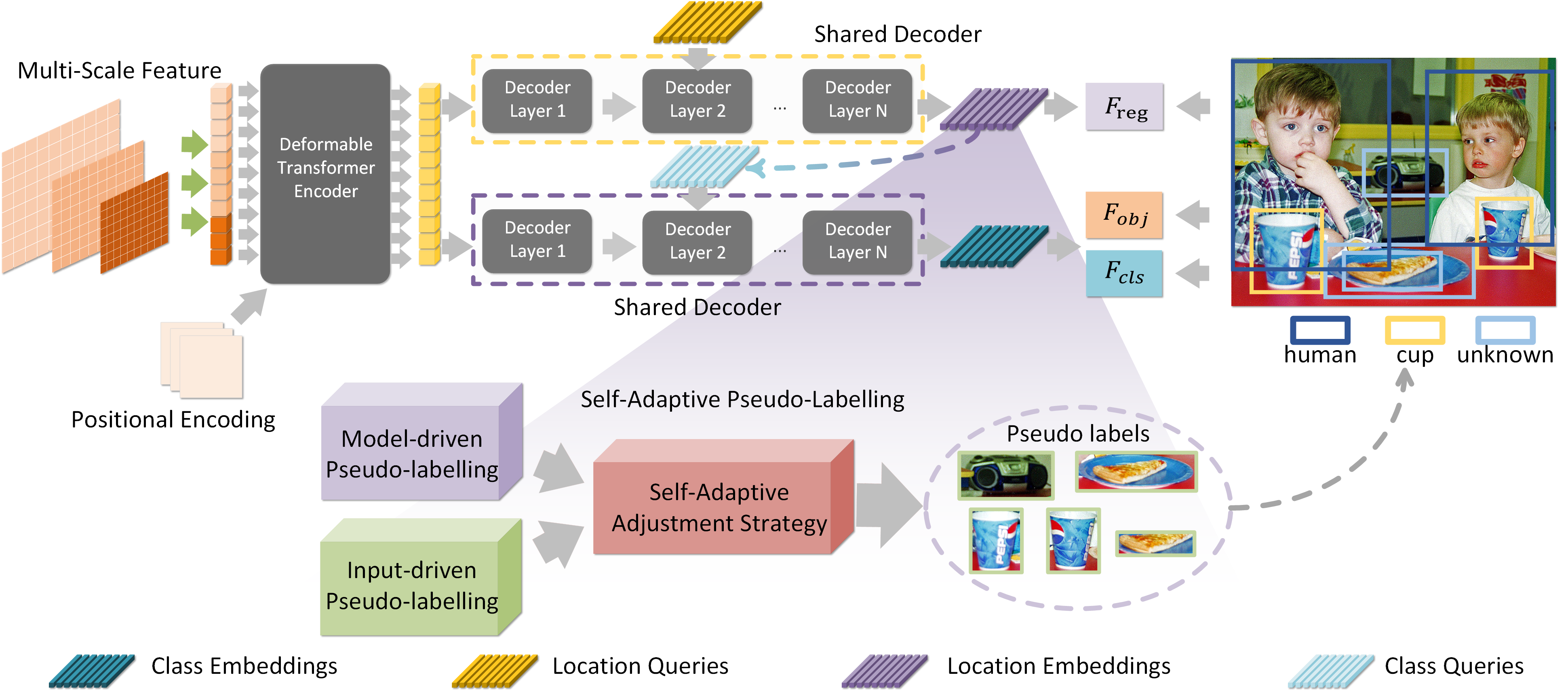}
  \caption{Overall Architecture  of proposed CAT framework. The proposed CAT consists of a multi-scale feature extractor, the shared transformer decoder, the regression prediction branch, and the self-adaptive pseudo-labelling. The multi-scale feature extractor comprises the mainstream feature extraction backbone and a deformable transformer encoder, for extracting multi-scale features. The shared transformer decoder is a deformable transformer decoder and decouples the localization and identification process in the cascade decoding way. The regression prediction branch contains the bounding box regression branch $F_{reg}$, novelty objectness branch $F_{obj}$, and novelty classification branch $F_{cls}$. While the novelty classification and objectness branches are single-layer feed-forward networks (FFN) and the regression branch is a 3-layer FFN. 
  }
  \label{fig:Figure2}
\end{figure*}
The Open-World Object Detection removes the artificial assumptions and restrictions in traditional object detection and makes object detection tasks more aligned with real life. It requires the trained model $\mathcal{M}_{t}$ not only to detect the previously encountered known classes $C$ but also to identify an unseen class instance as belonging to the unknown class. In addition, it requires the object detector to be capable of incremental update for new knowledge and this cycle continues over the detector’s lifespan. In incremental updating phase, the unknown instances identified by $\mathcal{M}_{t}$ are annotated manually, and along with their corresponding training examples, update $\mathcal{D}^{t}$ to $\mathcal{D}^{t+1}$ and $\mathcal{K}^{t}$ to $\mathcal{K}^{t+1}=\{1,2,\ldots, C,\ldots, C+\text{n}\}$, the model adds the $n$ new classes to known classes and updates itself to $\mathcal{M}_{t+1}$ without retraining from scratch on the whole dataset $\mathcal{D}^{t+1}$.

\section{Proposed method}

This section elaborates the proposed CAT in details. In Sec.\ref{3.1}, the overall architecture of CAT is described in detail.  We propose to decouple the decoding process of the detection transformer and propose the localization and identification cascade decoupled decoding manner in Sec.\ref{3.2}. A novel self-adaptive adjustment strategy for pseudo-labelling is proposed in Sec.\ref{3.3}. In Sec.\ref{3.4}, we illustrate the end-to-end training strategy of CAT.

\subsection{Overall Architecture}\label{3.1}
As shown in Figure.\ref{fig:Figure2}, for a given image $\mathcal{J} \in \mathbb{R}^{H \times W \times3}$, CAT uses a hierarchical feature extraction backbone to extract multi-scale features $\mathrm{Z}_{i} \in \mathbb{\mathbb { R }} ^{ \frac{\mathrm{H}}{4 \times i^{2}} \times \frac{w}{4 \times 2^{i}} \times 2^{i} C_{s}}, i=1,2,3$. The feature maps ${Z}_{i}$ are projected from dimension $C_s$ to dimension $C_d$ by using 1×1 convolution and concatenated to $N_s$ vectors with $C_d$ dimensions after flattening out.Afterwards, along with supplement positional encoding $P_n \in \mathbb{R}^{N_{s} \times C_{d}}$, the multi-scale features are sent into the deformable transformer encoder to encode semantic features. The encoded semantic features $M \in \mathbb{R}^{N_{s}\times c_{d}}$ are acquired and sent into the shared decoder together with a set of $N$ learnable location queries. Aided by interleaved cross-attention and self-attention modules, the shared decoder transforms the location queries $\mathcal{Q}_{\ \text {location}} \in \mathbb{R}^{N \times D}$ to a set of N location query embeddings $\mathcal{E}_{\ \text {location}} \in \mathbb{R}^{N \times D}$. The $\mathcal{E}_{\text {location}} $ are then input to the regression branch to locate N foreground bounding boxes containing the known classes and unknown classes. Meanwhile, the $\mathcal{E}_{\ \text {location}} $ are used as class queries and sent into the shared decoder together with the $M$ again. The shared decoder transforms the class queries to $N$ class query embeddings $\mathcal{E}_{\text {class}}$ that are corresponding to the location query embeddings. The $\mathcal{E}_{\text {class}}$ are then sent into the objectness and novelty classification branch to predict the objectness and category respectively. After selecting the unique queries that best match the known instances by a bipartite matching loss, the remaining queries are utilized to select the unknown category instances and generate pseudo labels by self-adaptive pseudo-labelling mechanism.
\subsection{The Cascade Decoupled Decoding Way}\label{3.2}

Detection transformer \cite{detr,ddetr,misra2021end,beal2020toward,li2022exploring,dai2021dynamic} leverages the object queries to detect object instances, where each object query represents an object instance. In the decoding stage, the object queries are updated to query embeddings by connecting object queries with semantic information from the encoded semantic features. The generated query embeddings couple the location and category information for both object localization and identification process simultaneously. For open-world object detection, the model requires detecting the known objects, localizing the unknown objects, and identifying them as the unknown class. 
Inspired by how people react to new scenarios \cite{CARRASCO20111484}, a cascade decoupled decoding manner is proposed to decode the encoded features in a cascade way.
We leverage the shared decoder to decode the encoded features twice. The first decoded embeddings are utilized to localize the foreground objects, while the second decoded embeddings are leveraged to identify the object categories and ``unknown''. The operation of localization and identification cascade decoding structure is expressed as follows:
\begin{equation}\label{eq5}
\mathcal{E}_{\text {Location }}=\mathcal{F}_{s}\left(\mathcal{F}_e(\varnothing(\mathcal{J}), P_n), \mathcal{Q}_{\ \text {Location}}, \mathcal{R}\right),
\end{equation}
\begin{equation}\label{eq6}
\mathcal{E}_{\text {Class }}=\mathcal{F}_{s}\left(\mathcal{F}_e(\varnothing(\mathcal{J}), P_n), \mathcal{E}_{\text {Location}}, \mathcal{R}\right).
\end{equation}
where $\mathcal{F}_{s}(\cdot)$ denotes the shared decoder. $\mathcal{F}_{e}(\cdot)$ is the encoder and $\varnothing(\cdot)$ is the backbone. $P_n$ stands for the positional encoding. $\mathcal{R}$ represents the reference points and $\mathcal{J}$ denotes the input image. In the cascade decoupled decoding phase, the location embeddings are used as class queries to generate class embeddings. Therefore, the localization process is not restricted by the category information, and the identification process can get help from the location knowledge in the cascade structure.

\subsection{Self-Adaptive Pseudo-labelling Mechanism}\label{3.3}
Pseudo labels play an important role in guiding models to detect unknown object instances, determining the upper learning limitation of the model. The existing methods \cite{ORE,owdetr} only use model-driven pseudo-labelling and do not take full advantage of the inputs' priori knowledge (light flow, textures, $etc$). The model-driven pseudo-labelling \cite{owdetr} makes the model’s learning get caught up in the knowledge of known objects, for the reason that the only source of knowledge for the model is known object instances. In addition, their fixed selection manner cannot guarantee the right learning direction for unknown objects. We propose to combine model-driven with input-driven pseudo-labelling \cite{edge,selective,Multiscale} for expanding the knowledge sources of the model. In the meanwhile, the pseudo-labels selection scheme should not be fixed, but be adapted as training and able to adjust itself when facing unexpected problems. 

\begin{algorithm}
        \caption{{\large C}OMPUTINGADAPTIVEWEIGHTS}  
        \label{alg:self-adaptive weight}
        \begin{algorithmic}[1] 
            \Require Loss Memory: $L_{m}$; Current Iteration: $t$; Positive Momentum Amplitude: $\large{\pi}_{pma}$; Negative Momentum Amplitude: $\large{\pi}_{nma}$; $T_{start}$: Start iteration; $T_b$: Weight updating cycle;
            Loss$ \leftarrow$ Compute using Equation.\ref{eq13}
            \Ensure self-adaptive weights $W_m{ }^t$ and $W_I{ }^t$
            \While{$train$}
                \If {$t \leq T_{start}$}
                    \State Initialise $W_m{ }^0 \leftarrow 0.8$ and $W_I{ }^0 \leftarrow 0.2$
                    \State Initialise $L_{m}$ using Equation.\ref{eq8}
                \Else
                    \State Update $L_{m}$ using Equation.\ref{eq8}
                    \If{$t \% T_{b}==0$}
                        \State Compute $\Delta l$ using $L_{m}$ and Equation.\ref{eq9} 
                        \State Compute $\Delta w$ using $\Delta l$ and Equation.\ref{eq11} 
                        \State Update $\mathcal{W}_m{ }^t$ and $\mathcal{W}_I{ }^t$ using Equation.\ref{eq12}
                    \EndIf
                \EndIf
            \EndWhile  
        \end{algorithmic}     
\end{algorithm} 

In this paper, inspired by \cite{pid}, a novel pseudo-labelling mechanism is proposed for self-adaptively combining model-driven and input-driven pseudo-labelling according to the situation faced by the model, where the attention-driven pseudo-labelling \cite{owdetr} is used as the model-driven pseudo-labelling and selective search \cite{selective} is selected as the input-driven pseudo-labelling. In the self-adaptive pseudo-labelling mechanism, the model-driven pseudo-labelling generates pseudo-labels' candidate boxes $P^{m}$ and the corresponding confidence $s_o$, and the input-driven pseudo-labelling generates pseudo-label candidate boxes $P^{I}$. The object confidence of generated pseudo labels is formulated as follows:
\begin{equation}\label{eq7}
\mathrm{S}_i=\left(norm\left(s_o\right)\right)^{\mathcal{W}_m} \cdot\left(\max _{1 \leq j \leq\left|\mathrm{P}^I\right|}\left(\operatorname{IOU}\left(P_j^I, P_i^m\right)\right)\right)^{\mathcal{W}_I},
\end{equation}
where IOU($\cdot$) (Intersection-over-union \cite{IoU}) is the most commonly used metric for comparing the similarity between two arbitrary shapes, $i$ denotes the index of the pseudo labels. $\mathcal{W}_m$ and $\mathcal{W}_I$ are the self-adaptive weights, which are controlled by the $Measurer$, $Sensor$ and $Adjuster$, as formulated below:
\begin{equation}
\mathcal{W}^t=Adjuster(\mathcal{W}^{t-1},\ Sensor(Measurer(L_m))),
\end{equation}
where $L_m$ represents the loss memory which is stored and updated in real time during model training. The formulation is illustrated in Equation.\ref{eq8}:
\begin{equation}
\label{eq8}
L_{m} = \text{DEQUE}({loss}_{t-1}, {loss}_{t-2}, \cdots, {loss}_{t-n}),
\end{equation}
where  $\text{DEQUE}$ is the sequence function, and $t$ is the current iteration. Considering the sensitivity of the model and the uneven quality of the data, we leverage $Measurer$ to obtain the trend of the losses $\Delta l$ for replacing the single loss. The formula is as follows:
\begin{equation}\label{eq9}
\begin{gathered}
Measurer(L_m) = \frac{\sum_{i=1}^n \alpha_i \cdot loss_{t-i}}{\sum_{j=n+1}^N \beta_j \cdot loss_{t-j}},\ \ \  n<N<T, 
\end{gathered}
\end{equation}
where $\alpha$ and $\beta$ denote the weighted average weights and they are the decreasing series of equal differences (\ie $\sum_{i=1}^n \alpha_i=\sum_{j=n+1}^N \beta_j=\frac{\alpha_i-\alpha_{i-1}}{\alpha_{i+1}-\alpha_i}=\frac{\beta_j-\beta_{j-1}}{\beta_{j+1}-\beta_j}=1$).
In the $Sensor$, the variable of the weight $\Delta w$ is acquired as follows:
\begin{equation}\label{eq11}
 Sensor(\Delta l) =\left\{\begin{array}{cl}
\large{\pi}_{nma} \cdot Sigmoid(\Delta l-1),  \Delta l>1, \\
-\large{\pi}_{pma} \cdot \Delta l, \Delta l \leq 1,
\end{array}\right.
\end{equation}
where $\large{\pi}_{pma}$ and $\large{\pi}_{nma}$ represents the positive and negative momentum amplitude (\ie the amplitude of incremental changing), respectively. In the $Adjuster$, we use Equation.\ref{eq12} to update the self-adaptive weight via a incremental way \cite{moco,ORE,mocov2}, for memory storage and enhancing the robustness.

\begin{equation}\label{eq12}
\left\{\begin{array}{l}
\mathcal{W}_m^{\ t} =\mathcal{W}_m^{\ t-1}+\Delta w \times \mathcal{W}_m^{\ t-1}, \\
\mathcal{W}_I^{\ t} =\mathcal{W}_I^{\ t-1}-\Delta w \times \mathcal{W}_I^{\ t-1}, \\
\mathcal{W}_m^{\ t}, \mathcal{W}_I^{\ t} = norm \left(\mathcal{W}_m^{\ t}, \mathcal{W}_I^{\ t}\right),
\end{array}\right.
\end{equation}
where $norm(\cdot)$ is the normalization operation. The update strategy for the weights during training is shown in Algorithm.\ref{alg:self-adaptive weight}.

\begin{figure*}[h]
  \centering
  \includegraphics[width = \linewidth]{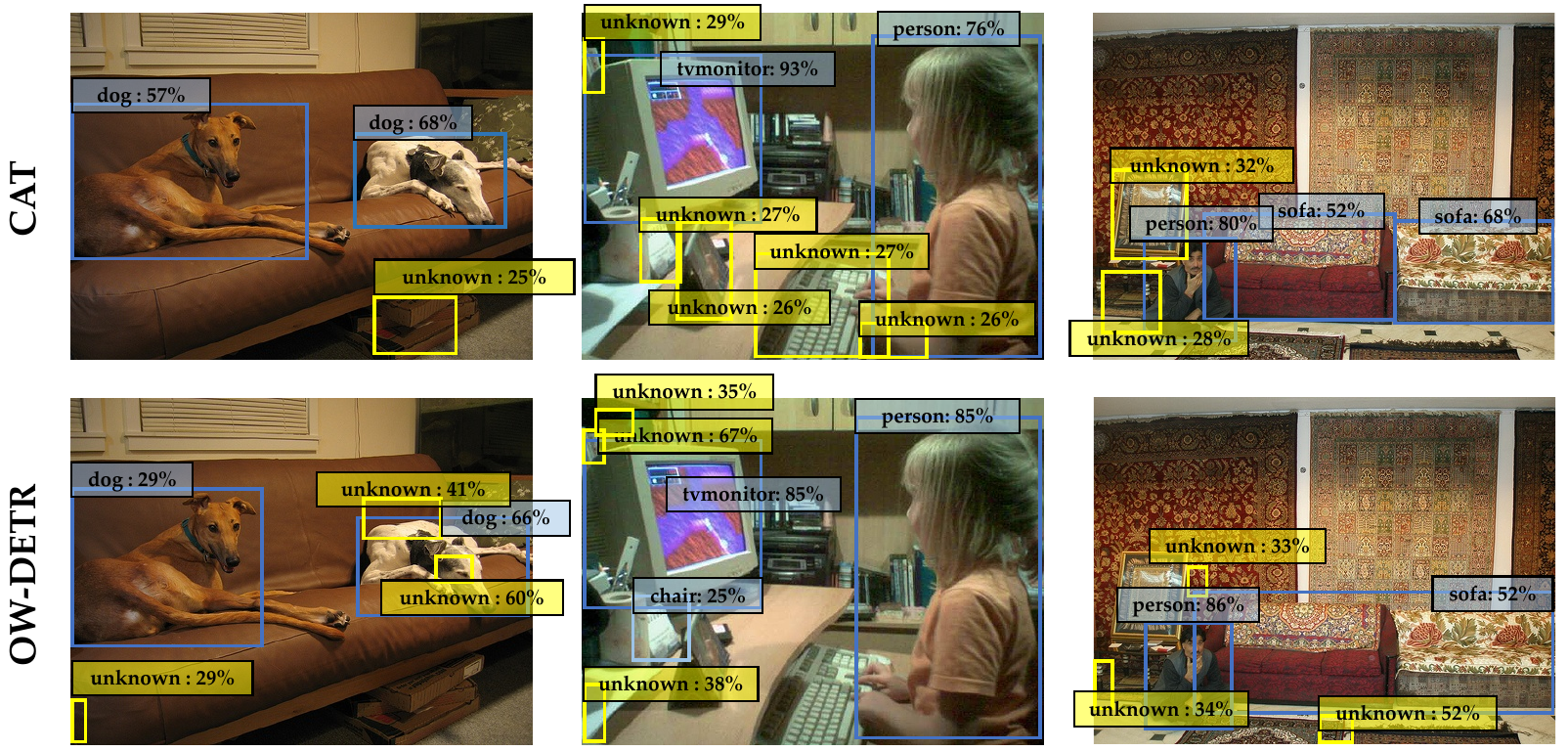}
  \caption{\textbf{Comparison of qualitative results between CAT and OW-DETR}. Detections of CAT \ (top row) and OW-DETR (bottom row) are displayed, with  \textcolor[rgb]{0.1,0.8,0.9}{Blue} - known and  \textcolor[RGB]{255,215,0}{Yellow} - unknown object detections. CAT detected more unknown objects than OW-DETR. In the left column, OW-DETR identifies the background and known objects as unknowns and the real unknown object (\textit{carton}) as the background, and our model accurately identifies the \textit{carton} as the unknown object. In the middle column, OW-DETR identifies the two \textit{calendars} as the \textit{chair} and the background, respectively, and the \textit{keyboard} as the background, and our CAT accurately identifies them as unknown objects. The right column shows that OW-DETR not only does not detect the unknown object (\textit{frame}) but also identifies two known objects (\textit{sofa}) as one. Our model accurately identifies the frame as an unknown object and also accurately identifies the two \textit{sofas}.}
  \label{fig:Figure3}
\end{figure*}

\subsection{Training and Inference}\label{3.4}
Our CAT is trained end-to-end using the following joint loss formulation:
\begin{equation}\label{eq13}
L=L_{localization}+L_{identification}+L_{objectness}, 
\end{equation}
where $L_{localization}$, $L_{identification}$ and $L_{objectness}$ denotes the loss terms for foreground localization, novelty identification and object scoring, respectively. When a set of new categories are introduced at each episode, we employ an exemplar replay based finetuning to alleviate catastrophic forgetting of learned classes and then finetune the model using a balanced set of exemplars stored for each known class. The bounding boxes and categories predictions of the known and $top$-k unknown objects are simultaneous used during evaluation.

\section{Experiments}

\begin{table*}[htbp]
\renewcommand\arraystretch{1.25}
\centering
\caption{\textbf{State-of-the-art comparison on OWOD split}. The comparison is shown in terms of U-Recall and known class mAP. U-Recall measures the ability of the model to retrieve unknown object instances for OWOD problem. For a fair comparison, we compare with the recently introduced methods and ORE not employing EBUI. The CAT achieves improved all metrics over the existing works across all tasks, demonstrating our model’s effectiveness for OWOD problem. U-Recall cannot be computed in Task 4 due to the absence of unknown test annotations, for the reason that all 80 classes are known.} \label{table1}
\resizebox{\textwidth}{!}{
\begin{tabular}{l| cc| cccc |cccc |ccc}\toprule
\multicolumn{1}{c|}{\textbf{Task IDs $\rightarrow$}}& \multicolumn{2}{c|}{\textbf{Task 1}}& \multicolumn{4}{c|}{\textbf{Task 2}} & \multicolumn{4}{c|}{\textbf{Task 3}}  & \multicolumn{3}{c}{\textbf{Task 4}} \\\midrule
\multicolumn{1}{c|}{}   & \multicolumn{1}{c}{ \cellcolor[HTML]{fcfce0}Unknown}  & \cellcolor[HTML]{e4e5fa}Known &  \multicolumn{1}{c}{ \cellcolor[HTML]{fcfce0}Unknown} & \multicolumn{3}{c|}{\cellcolor[HTML]{e4e5fa}Known} & \multicolumn{1}{c}{ \cellcolor[HTML]{fcfce0}Unknown} & \multicolumn{3}{c|}{\cellcolor[HTML]{e4e5fa}Known}  &\multicolumn{3}{c}{\cellcolor[HTML]{e4e5fa}Known}  \\
\multicolumn{1}{l|}{}    & \cellcolor[HTML]{fcfce0} Recall&  \cellcolor[HTML]{e4e5fa}mAP($\uparrow$) & \cellcolor[HTML]{fcfce0}Recall& \multicolumn{3}{c|}{\cellcolor[HTML]{e4e5fa}mAP($\uparrow$)}&\cellcolor[HTML]{fcfce0}Recall  &\multicolumn{3}{c|}{\cellcolor[HTML]{e4e5fa}mAP($\uparrow$)}   &  \multicolumn{3}{c}{\cellcolor[HTML]{e4e5fa}mAP($\uparrow$)}\\

\multicolumn{1}{c|}{\multirow{-3}{*}{Metrics $\rightarrow$}}   & \multirow{-1}{*}{\cellcolor[HTML]{fcfce0}($\uparrow$)}& \cellcolor[HTML]{e4e5fa}Current & \multirow{-1}{*}{\cellcolor[HTML]{fcfce0}($\uparrow$)}   &\cellcolor[HTML]{e4e5fa}Previously & \cellcolor[HTML]{e4e5fa}Current &\cellcolor[HTML]{e4e5fa}Both & \multirow{-1}{*}{\cellcolor[HTML]{fcfce0}($\uparrow$)}  &\cellcolor[HTML]{e4e5fa}Previously & \cellcolor[HTML]{e4e5fa}Current &\cellcolor[HTML]{e4e5fa}Both &\cellcolor[HTML]{e4e5fa}Previously & \cellcolor[HTML]{e4e5fa}Current &\cellcolor[HTML]{e4e5fa}Both \\ \midrule

UC-OWOD\cite{UC-OWOD} &2.4  & 50.7 & 3.4  & 33.1& 30.5 & 31.8 &8.7  &28.8&16.3&24.6&25.6&15.9&23.2 \\
ORE-EBUI\cite{ORE}& 4.9  & 56.0   &2.9  & 52.7 & 26.0  & 39.4  & 3.9   & 38.2  & 12.7 & 29.7  & 29.6   & 12.4  & 25.3  \\
OW-DETR\cite{owdetr}  &7.5 &  59.2 & 6.2 & 53.6  & \textbf{33.5}   & 42.9  & 5.7  & 38.3 & 15.8 & 30.8  & 31.4 & \textbf{17.1} & 27.8  \\ 
OCPL\cite{OCPL} &8.3  & 56.6 &7.7  & 50.6    & 27.5 & 39.1 &11.9  &38.7&14.7&30.7&30.7&14.4 &26.7\\
2B-OCD\cite{two-branch} &12.1  & 56.4 & 9.4  & 51.6& 25.3 & 38.5 &11.6  & 37.2 & 13.2 & 29.2&30.0&13.3&25.8 \\
 \multirow{1}{*}{\textbf{Ours: CAT}}   &\textbf{23.7}    &\textbf{60.0} & \textbf{19.1} & \textbf{55.5} & 32.7 & \textbf{44.1}  & \textbf{24.4}&  \textbf{42.8} & \textbf{18.7}  & \textbf{34.8} &\textbf{34.4} & 16.6  &\textbf{ 29.9}\\ 

\bottomrule 
\end{tabular}}
\end{table*}

\begin{table*}[htbp]
\renewcommand\arraystretch{1.25}
\centering
\caption{\textbf{State-of-the-art comparison on MS-COCO split}. The comparison is shown in terms of U-Recall and mAP. Although the MS-COCO split is more challenging, our model gets a more significant improvement on this in comparison to ORE and OW-DETR. The significant metric improvements demonstrate that our CAT has the ability to retrieve new knowledge beyond the range of closed set and would not be limited by category knowledge of existing objects.}\label{table2}
\resizebox{\textwidth}{!}{
\begin{tabular}{l| cc| cccc |cccc |ccc}\toprule
\multicolumn{1}{c|}{\textbf{Task IDs $\rightarrow$}}& \multicolumn{2}{c|}{\textbf{Task 1}}& \multicolumn{4}{c|}{\textbf{Task 2}} & \multicolumn{4}{c|}{\textbf{Task 3}}  & \multicolumn{3}{c}{\textbf{Task 4}} \\\midrule
\multicolumn{1}{c|}{}   & \multicolumn{1}{c}{ \cellcolor[HTML]{fcfce0}Unknown}  & \cellcolor[HTML]{e4e5fa}Known &  \multicolumn{1}{c}{ \cellcolor[HTML]{fcfce0}Unknown} & \multicolumn{3}{c|}{\cellcolor[HTML]{e4e5fa}Known} & \multicolumn{1}{c}{ \cellcolor[HTML]{fcfce0}Unknown} & \multicolumn{3}{c|}{\cellcolor[HTML]{e4e5fa}Known}  &\multicolumn{3}{c}{\cellcolor[HTML]{e4e5fa}Known}  \\
\multicolumn{1}{l|}{}    & \cellcolor[HTML]{fcfce0} Recall&  \cellcolor[HTML]{e4e5fa}mAP($\uparrow$) & \cellcolor[HTML]{fcfce0}Recall& \multicolumn{3}{c|}{\cellcolor[HTML]{e4e5fa}mAP($\uparrow$)}&\cellcolor[HTML]{fcfce0}Recall  &\multicolumn{3}{c|}{\cellcolor[HTML]{e4e5fa}mAP($\uparrow$)}   &  \multicolumn{3}{c}{\cellcolor[HTML]{e4e5fa}mAP($\uparrow$)}\\

\multicolumn{1}{c|}{\multirow{-3}{*}{Metrics $\rightarrow$}}   & \multirow{-1}{*}{\cellcolor[HTML]{fcfce0}($\uparrow$)}& \cellcolor[HTML]{e4e5fa}Current & \multirow{-1}{*}{\cellcolor[HTML]{fcfce0}($\uparrow$)}   &\cellcolor[HTML]{e4e5fa}Previously & \cellcolor[HTML]{e4e5fa}Current &\cellcolor[HTML]{e4e5fa}Both & \multirow{-1}{*}{\cellcolor[HTML]{fcfce0}($\uparrow$)}  &\cellcolor[HTML]{e4e5fa}Previously & \cellcolor[HTML]{e4e5fa}Current &\cellcolor[HTML]{e4e5fa}Both &\cellcolor[HTML]{e4e5fa}Previously & \cellcolor[HTML]{e4e5fa}Current &\cellcolor[HTML]{e4e5fa}Both \\ \midrule

ORE-EBUI\cite{ORE}& 1.5 & 61.4   & 3.9& 56.5 & 26.1  & 40.6  &3.6   & 38.7  & 23.7 & 33.7   & 33.6   & 26.3  & 31.8  \\
OW-DETR\cite{owdetr}  & 5.7 &  71.5 & 6.2 & 62.8  & 27.5   & 43.8  & 6.9& 45.2 & 24.9 & 38.5  & 38.2 & 28.1 & 33.1  \\ 

 \multirow{1}{*}{\textbf{Ours: CAT}}   &\textbf{24.0}  &\textbf{74.2} & \textbf{23.0} &\textbf{67.6} & \textbf{35.5}  & \textbf{50.7}  & \textbf{24.6}&  \textbf{51.2} &\textbf{32.6} &\textbf{ 45.0}& \textbf{45.4} &\textbf{35.1} &\textbf{42.8} \\ 
\bottomrule 
\end{tabular}}
\end{table*}

\begin{table*}[htbp]
\renewcommand\arraystretch{1.25}
\centering
\caption{\textbf{Component ablation experiment}. The comparison is shown in terms of known class average precision (mAP) and unknown class recall (U-Recall). \textbf{CAT}\texttt{-Cddw} is our model without the cascade decoupled decoding way. \textbf{CAT}\texttt{-Sam} is our model without the self-adaptive manner but with the prior from the selective search. We also include the performance of deformable DETR and an upper bound (D-DETR trained with ground-truth unknown class annotations) as reported by OW-DETR \cite{owdetr}.}\label{table3}
\resizebox{\textwidth}{!}{
\begin{tabular}{l| cc| cccc |cccc |ccc}\toprule
\multicolumn{1}{c|}{\textbf{Task IDs $\rightarrow$}}& \multicolumn{2}{c|}{\textbf{Task }1}& \multicolumn{4}{c|}{\textbf{Task 2}} & \multicolumn{4}{c|}{\textbf{Task 3}}  & \multicolumn{3}{c}{\textbf{Task 4}} \\\midrule
\multicolumn{1}{c|}{}   & \multicolumn{1}{c}{ \cellcolor[HTML]{fcfce0}Unknown}  & \cellcolor[HTML]{e4e5fa}Known &  \multicolumn{1}{c}{ \cellcolor[HTML]{fcfce0}Unknown} & \multicolumn{3}{c|}{\cellcolor[HTML]{e4e5fa}Known} & \multicolumn{1}{c}{ \cellcolor[HTML]{fcfce0}Unknown} & \multicolumn{3}{c|}{\cellcolor[HTML]{e4e5fa}Known}  &\multicolumn{3}{c}{\cellcolor[HTML]{e4e5fa}Known}  \\
\multicolumn{1}{l|}{}    & \cellcolor[HTML]{fcfce0} Recall&  \cellcolor[HTML]{e4e5fa}mAP($\uparrow$) & \cellcolor[HTML]{fcfce0}Recall& \multicolumn{3}{c|}{\cellcolor[HTML]{e4e5fa}mAP($\uparrow$)}&\cellcolor[HTML]{fcfce0}Recall  &\multicolumn{3}{c|}{\cellcolor[HTML]{e4e5fa}mAP($\uparrow$)}   &  \multicolumn{3}{c}{\cellcolor[HTML]{e4e5fa}mAP($\uparrow$)}\\

\multicolumn{1}{c|}{\multirow{-3}{*}{Metrics $\rightarrow$}}   & \multirow{-1}{*}{\cellcolor[HTML]{fcfce0}($\uparrow$)}& \cellcolor[HTML]{e4e5fa}Current & \multirow{-1}{*}{\cellcolor[HTML]{fcfce0}($\uparrow$)}   &\cellcolor[HTML]{e4e5fa}Previously & \cellcolor[HTML]{e4e5fa}Current &\cellcolor[HTML]{e4e5fa}Both & \multirow{-1}{*}{\cellcolor[HTML]{fcfce0}($\uparrow$)}  &\cellcolor[HTML]{e4e5fa}Previously & \cellcolor[HTML]{e4e5fa}Current &\cellcolor[HTML]{e4e5fa}Both &\cellcolor[HTML]{e4e5fa}Previously & \cellcolor[HTML]{e4e5fa}Current &\cellcolor[HTML]{e4e5fa}Both \\ \midrule
Upper Bound & 31.6  & 62.5  & 40.5 & 55.8 & 38.1 & 46.9 & 42.6 & 42.4 & 29.3 & 33.9 & 35.6 & 23.1 & 32.5 \\ 
D-DETR \cite{ddetr} & -&60.3  & - & 54.5 & 34.4 & 44.7  & - & 40.0 & 17.7 & 33.3 & 32.5 & 20.0 & 29.4 \\ 
\midrule
\midrule
\textbf{CAT} -- \texttt{Cddw}  &19.1&59.3&18.6&52.8&30.2&41.5&21.0&41.0&17.6&33.0&32.6&15.8&27.9    \\
\textbf{CAT} -- \texttt{Sam} &19.1&59.7&16.9&54.8&32.4&43.6&18.6&42.1&19.3&34.5&34.0&16.0&29.5  \\
Final:\textbf{CAT}  &23.7&60.0&19.1&55.5&32.7&44.1&24.4&42.8&18.7&34.8&34.4&16.6&29.9   \\
\bottomrule 
\end{tabular}}
\label{table5}
\end{table*}

\subsection{Datasets and Metrics} 
The experiments are implemented on two mainstream splits of MS-COCO \cite{mscoco} and Pascal VOC \cite{voc} dataset. We group the classes into a set of nonoverlapping tasks $\left\{T^1, \ldots, T^t, \ldots\right\}$. The class in task $T^c$ only appears in tasks where $t \geq c$. In task $T^c$, classes encountered in $\left\{T^c: c \leq t\right\}$ and $\left\{T^c: c>t\right\}$ are considered as known and unknown classes, respectively. \par
\noindent \textbf{OWOD SPLIT} \cite{ORE} spilts the 80 classes of MS-COCO into 4 tasks and selects training set for each task from the MS-COCO and Pascal VOC training set images. Pascal VOC testing and MS-COCO validation set are used for evaluation. \par
\noindent \textbf{MS-COCO SPLIT} \cite{owdetr} mitigates data leakage across tasks in \cite{ORE} and is more challenging. The training and testing data are selected from MS-COCO.\par
\noindent \textbf{Metrics:} 
Following the most commonly used evaluation metric for object detection, we use mean average precision (mAP) to evaluate the known objects. Inspired by \cite{ORE,recall,owdetr,wi,A-OSE}, U-Recall is used as main metric for unknown objects. U-Recall measures the ability of the model to retrieve unknown object instances for OWOD problem. 

\subsection{Implementation Details}
The multi-scale feature extractor consists of a Resnet-50\cite{resnet50} pretrained on ImageNet\cite{imagenet} in a self-supervised\cite{self} manner and a deformable transformer encoder whose number of layer is set to 6. For the shared decoder, we use a deformable transformer decoder and the numbder of layer is set to 6, too. We set the number of queries $M=100$, the dimension of the embeddings $D=256$ and the number of pseudo-labels $k=5$. During inference, $top$-50 high scoring detections are used for evaluation for per image.

\subsection{Comparison With State-of-the-art Methods}\label{4.3}
For a fair comparison, we compare CAT with ORE \cite{ORE} without the energy-based unknown identifier (EBUI) that relies on held-out validation data with weak unknown object supervision and other SOTA methods \cite{owdetr,OCPL,two-branch,UC-OWOD} to demonstrate the effectiveness of our method for OWOD problem. We present the comparison in terms of known class mAP and unknown class recall where U-Recall cannot be computed in Task 4 due to the absence of unknown test annotations, for the reason that all 80 classes are known.  
\par
\noindent \textbf{OWOD SPLIT:}
The results compared with the state-of-the-art methods on OWOD split for OWOD problem are shown in Table.\ref{table1}. Benefiting from the cascade decoupled decoding manner and the self-adaptive pseudo-labelling mechanism, the ability of CAT to detect unknown objects goes substantially beyond the existing models. Compared with 2B-OCD's \cite{two-branch} U-Recall of 12.1, 9.4 and 11.6 on Task 1, 2 and 3, our CAT achieves 23.7, 19.1 and 24.4 in the corresponding tasks, achieving significant absolute gains up to 12.8\%. The ability to detect known objects and alleviate catastrophic forgetting of previous knowledge gains an improved performance with significant gains, achieving significant absolute gains up to 4.7\% beyond OW-DETR \cite{owdetr}. This demonstrates the significant performance of the cascade decoding manner. 
\par
\noindent \textbf{MS-COCO SPLIT:}
We report the results on MS-COCO split in Table.\ref{table2}. MS-COCO split mitigates data leakage across tasks and assign more data to each Task, while CAT receives a more significant boost compared with OWOD split. Compared with OW-DETR's U-Recall of 5.7, 6.2 and 6.9 on Task 1, 2 and 3, our CAT achieves 24.0, 23.0 and 24.6 in the corresponding tasks, achieving significant absolute gains up to 18.3\%. Furthermore, the performance on detecting known objects achieves significant absolute gains up to 9.7\%. This demonstrates that our CAT has the more powerful ability to retrieve new knowledge and detect the known objects when faced with more difficult tasks.
\par
\noindent \textbf{Qualitative Results:}
We report qualitative results in Figure.\ref{fig:Figure3}. We show the detection results of CAT \ (top row) and OW-DETR (bottom row), with  \textcolor[rgb]{0.1,0.8,0.9}{Blue} - known objects and \textcolor[RGB]{255,215,0}{Yellow} - unknown objects. It is easy to see that CAT could detect more unknown objects. In the left column, OW-DETR identifies the background and known objects (\textit{dog}) as unknowns and the real unknown object (\textit{carton}) as the background, while our model accurately identifies the \textit{carton} as the unknown object. As shown in the middle column, OW-DETR identifies the two \textit{calendars} as the \textit{chair} and the background, respectively, and the \textit{keyboard} as the background, while our CAT accurately identifies them as unknown objects. The right column shows that OW-DETR not only does not detect the unknown object (\textit{frame}) but also identifies two known objects (\textit{sofa}) as one, while our model accurately identifies the frame as an unknown object and accurately identifies the two \textit{sofas}.

\subsection{Ablation Study}\label{4.4}
 We conduct abundant ablative experiments to verify the effectiveness of CAT's components on the OWOD split. Furthermore, we demonstrate the effectiveness of our model for incremental object detection and open-set detection.\par
\noindent \textbf{Ablating Components:} 
To study the contribution of each component, we design ablation experiments in Table.\ref{table5}. In comparison to the Final CAT, removing the cascade decoupled decoding manner \textbf{CAT}\texttt{-Cddw} reduces the performance on retrieving unknown objects and detecting known objects, achieving significant absolute gains down to 
4.6, 0.5 and 3.4 points in Task 1,2,3 for U-Recall and the mAP for known objects is reduced by 0.7, 2.6, 1.8, and 2.0 in Task 1,2,3,4. The results demonstrate that the cascade decoupled decoding manner is better for the open-world object detection which contains the unknown objects and improves the ability of CAT to retrieve unknown objects and detect known objects. To ablate the self-adaptive manner component, we remove the self-adaptive manner from CAT and hold the prior from selective search. Compared with CAT, removing the self-adaptive manner \textbf{CAT}\texttt{-Sam} significantly reduces the performance on detecting unknown objects, achieving significant absolute gains down to 4.6, 2.2, and 5.8 points in Task 1,2,3 respectively. The results demonstrate that the self-adaptive manner could efficiently combine the input and model-driven pseudo-labelling mechanism, improving the CAT's ability to explore unknown objects. Thus, each component has a critical role to play in open-World object detection. \par
\par

\noindent \textbf{Incremental Object Detection:}
To intuitively present our CAT's ability for detecting object instances, we compare it to \cite{iod,fasteriod,ORE,owdetr} on the incremental object detection (IOD) task. We evaluate the experiments on three standard settings, where a
group of classes (10, 5 and last class) are introduced incrementally to a detector trained on the remaining classes (10, 15 and 19), based on PASCAL VOC 2007 dataset \cite{voc}. As the results shown in Table.\ref{table4}, CAT outperforms the existing method in a great migration on all three settings, indicating the power of localization and identification cascade detection transformer for IOD.\par



\begin{table*}[htbp]
\renewcommand\arraystretch{1.25}
\caption{\textbf{Performance comparison on incremental object detection task.} Evaluation is performed on three standard settings, where a group of classes (10, 5 and last class) are introduced incrementally to a detector trained on the remaining classes (10,15 and 19). Our CAT performs favorably against existing approaches on all three settings, illustrating the power of localization identification cascade detection transformer for incremental objection detection.} 
\resizebox{\textwidth}{!}{
\begin{tabular}{lccccccccccccccccccccc}
\toprule
\color[HTML]{009901} \textbf{10 + 10 setting}  & aero & cycle & bird & boat & bottle & bus  & car  & cat  & chair & cow  & table  & dog  & horse  & bike   & person & plant  & sheep  & sofa   & train  & tv & mAP  \\
\midrule
ILOD\cite{iod} & 69.9 & 70.4  & 69.4 & 54.3 & 48   & 68.7 & 78.9 & 68.4 & 45.5  & 58.1 & \cellcolor[HTML]{FAF4E2}59.7&\cellcolor[HTML]{FAF4E2}72.7 &\cellcolor[HTML]{FAF4E2}73.5 &\cellcolor[HTML]{FAF4E2}73.2 &\cellcolor[HTML]{FAF4E2}66.3 &\cellcolor[HTML]{FAF4E2}29.5 &\cellcolor[HTML]{FAF4E2}63.4 &\cellcolor[HTML]{FAF4E2}61.6 &\cellcolor[HTML]{FAF4E2}69.3 &\cellcolor[HTML]{FAF4E2}62.2 & 63.2 \\
Faster ILOD\cite{fasteriod}  & 72.8 & 75.7  &  71.2  &  60.5  &  61.7  &  70.4  &  83.3  &  76.6  &  53.1  &  72.3  &\cellcolor[HTML]{FAF4E2} 36.7  &\cellcolor[HTML]{FAF4E2} 70.9  &\cellcolor[HTML]{FAF4E2} 66.8  &\cellcolor[HTML]{FAF4E2} 67.6  &\cellcolor[HTML]{FAF4E2} 66.1  &\cellcolor[HTML]{FAF4E2}24.7  &\cellcolor[HTML]{FAF4E2} 63.1  &\cellcolor[HTML]{FAF4E2} 48.1  &\cellcolor[HTML]{FAF4E2} 57.1  &\cellcolor[HTML]{FAF4E2} 43.6  & 62.1 \\
ORE-(CC+EBUI)\cite{ORE}  &53.3& 69.2& 62.4& 51.8& 52.9& 73.6& 83.7& 71.7& 42.8& 66.8& \cellcolor[HTML]{FAF4E2}46.8&\cellcolor[HTML]{FAF4E2}59.9&\cellcolor[HTML]{FAF4E2}65.5&\cellcolor[HTML]{FAF4E2}66.1&\cellcolor[HTML]{FAF4E2}68.6&\cellcolor[HTML]{FAF4E2}29.8&\cellcolor[HTML]{FAF4E2}55.1&\cellcolor[HTML]{FAF4E2}51.6&\cellcolor[HTML]{FAF4E2}65.3&\cellcolor[HTML]{FAF4E2}51.5& 59.4  \\
ORE-EBUI\cite{ORE} &63.5& 70.9& 58.9& 42.9& 34.1& 76.2& 80.7& 76.3& 34.1& 66.1&\cellcolor[HTML]{FAF4E2}56.1&\cellcolor[HTML]{FAF4E2}70.4&\cellcolor[HTML]{FAF4E2}80.2&\cellcolor[HTML]{FAF4E2}72.3&\cellcolor[HTML]{FAF4E2}81.8&\cellcolor[HTML]{FAF4E2}42.7&\cellcolor[HTML]{FAF4E2}71.6&\cellcolor[HTML]{FAF4E2}68.1&\cellcolor[HTML]{FAF4E2}77&\cellcolor[HTML]{FAF4E2}67.7& 64.5  \\

OW-DETR\cite{owdetr} &61.8 &69.1 &67.8& 45.8& 47.3& 78.3& 78.4& 78.6& 36.2& 71.5&\cellcolor[HTML]{FAF4E2} 57.5&\cellcolor[HTML]{FAF4E2} 75.3&\cellcolor[HTML]{FAF4E2} 76.2& \cellcolor[HTML]{FAF4E2}77.4& \cellcolor[HTML]{FAF4E2}79.5& \cellcolor[HTML]{FAF4E2}40.1& \cellcolor[HTML]{FAF4E2}66.8& \cellcolor[HTML]{FAF4E2}66.3& \cellcolor[HTML]{FAF4E2}75.6& \cellcolor[HTML]{FAF4E2}64.1& 65.7  \\\midrule
\textbf{Ours: CAT} & 76.5&75.7&67.0&51.0&62.4&73.2&82.3&83.7&42.7&64.4   &\cellcolor[HTML]{FAF4E2}56.8   &\cellcolor[HTML]{FAF4E2}74.1   &\cellcolor[HTML]{FAF4E2}75.8   &\cellcolor[HTML]{FAF4E2}79.2   &\cellcolor[HTML]{FAF4E2}78.1   &\cellcolor[HTML]{FAF4E2}39.9   &\cellcolor[HTML]{FAF4E2}65.1   &\cellcolor[HTML]{FAF4E2}59.6   &\cellcolor[HTML]{FAF4E2}78.4   &\cellcolor[HTML]{FAF4E2}67.4   &67.7 \\\midrule\midrule
\color[HTML]{009901} \textbf{15 + 5 setting}& aero & cycle & bird & boat & bottle & bus  & car  & cat  & chair & cow  & table  & dog  & horse  & bike & person & plant  & sheep  & sofa & train  & tv & mAP  \\\midrule
ILOD\cite{iod} & 70.5 & 79.2 &68.8& 59.1& 53.2& 75.4& 79.4& 78.8& 46.6& 59.4& 59 &75.8& 71.8& 78.6& 69.6&\cellcolor[HTML]{FAF4E2}33.7&\cellcolor[HTML]{FAF4E2}61.5&\cellcolor[HTML]{FAF4E2}63.1 &\cellcolor[HTML]{FAF4E2}71.7&\cellcolor[HTML]{FAF4E2}62.2& 65.8 \\
Faster ILOD\cite{fasteriod}  &66.5& 78.1& 71.8& 54.6& 61.4& 68.4& 82.6& 82.7& 52.1& 74.3& 63.1& 78.6& 80.5& 78.4& 80.4&\cellcolor[HTML]{FAF4E2}36.7&\cellcolor[HTML]{FAF4E2}61.7&\cellcolor[HTML]{FAF4E2}59.3&\cellcolor[HTML]{FAF4E2}67.9&\cellcolor[HTML]{FAF4E2}59.1 &67.9 \\
ORE-(CC+EBUI)\cite{ORE}  &65.1& 74.6& 57.9& 39.5 &36.7& 75.1& 80 &73.3& 37.1& 69.8& 48.8& 69 &77.5& 72.8& 76.5&\cellcolor[HTML]{FAF4E2}34.4&\cellcolor[HTML]{FAF4E2}62.6&\cellcolor[HTML]{FAF4E2}56.5&\cellcolor[HTML]{FAF4E2}80.3&\cellcolor[HTML]{FAF4E2}65.7& 62.6 \\
ORE-EBUI\cite{ORE} &75.4& 81& 67.1& 51.9& 55.7& 77.2& 85.6& 81.7& 46.1& 76.2& 55.4& 76.7& 86.2& 78.5& 82.1&\cellcolor[HTML]{FAF4E2}32.8&\cellcolor[HTML]{FAF4E2}63.6&\cellcolor[HTML]{FAF4E2}54.7&\cellcolor[HTML]{FAF4E2}77.7&\cellcolor[HTML]{FAF4E2}64.6 &68.5 \\
OW-DETR\cite{owdetr} &77.1 &76.5 &69.2 &51.3& 61.3 &79.8& 84.2& 81.0& 49.7 &79.6& 58.1& 79.0& 83.1 &67.8& 85.4& \cellcolor[HTML]{FAF4E2}33.2& \cellcolor[HTML]{FAF4E2}65.1 &\cellcolor[HTML]{FAF4E2}62.0 &\cellcolor[HTML]{FAF4E2}73.9& \cellcolor[HTML]{FAF4E2}65.0& 69.4\\\midrule
\textbf{Ours: CAT} &75.3&81.0&84.4&64.5&56.6&74.4&84.1&86.6&53.0&70.1&72.4&83.4&85.5&81.6&81.0&\cellcolor[HTML]{FAF4E2}32.0&\cellcolor[HTML]{FAF4E2}58.6&\cellcolor[HTML]{FAF4E2}60.7&\cellcolor[HTML]{FAF4E2}81.6&\cellcolor[HTML]{FAF4E2}63.5 &72.2 \\\midrule\midrule
\color[HTML]{009901} \textbf{19 + 1 setting}  & aero & cycle & bird & boat & bottle & bus  & car  & cat  & chair & cow  & table  & dog  & horse  & bike   & person & plant  & sheep  & sofa   & train  & tv & mAP  \\\midrule
ILOD\cite{iod} & 69.4&79.3&69.5&57.4&45.4&78.4&79.1&80.5&45.7&76.3&64.8&77.2&80.8&77.5&70.1&42.3&67.5&64.4&76.7&\cellcolor[HTML]{FAF4E2}62.7&68.2  \\
Faster ILOD\cite{fasteriod}   &64.2&74.7&73.2&55.5&53.7&70.8&82.9&82.6&51.6&79.7&58.7&78.8&81.8&75.3&77.4&43.1&73.8&61.7&69.8&\cellcolor[HTML]{FAF4E2}61.1&68.5  \\
ORE-(CC+EBUI)\cite{ORE}  &60.7&78.6&61.8&45&43.2&75.1&82.5&75.5&42.4&75.1&56.7&72.9&80.8&75.4&77.7&37.8&72.3&64.5&70.7&\cellcolor[HTML]{FAF4E2}49.9&64.9 \\
ORE-EBUI\cite{ORE}   &67.3&76.8&60&48.4&58.8&81.1&86.5&75.8&41.5&79.6&54.6&72.8&85.9&81.7&82.4&44.8&75.8&68.2&75.7&\cellcolor[HTML]{FAF4E2}60.1&68.8\\

OW-DETR\cite{owdetr} &70.5 &77.2 &73.8& 54.0& 55.6& 79.0 &80.8 &80.6 &43.2& 80.4 &53.5 &77.5 &89.5 &82.0& 74.7 &43.3 &71.9 &66.6& 79.4 &\cellcolor[HTML]{FAF4E2}62.0&70.2  \\\midrule
\textbf{Ours: CAT} & 86.0&85.8&78.8&65.3&61.3&71.4&84.8&84.8&52.9&78.4&71.6&82.7&83.8&81.2&80.7&43.7&75.9&58.5&85.2&\cellcolor[HTML]{FAF4E2}61.1 &73.8  \\ \bottomrule
\end{tabular}
}
\label{table4}
\end{table*}
\noindent \textbf{Open-set Detection Comparison:} To further demonstrate CAT's ability to handle unknown instances in open-set data, we follow the same evaluation protocol as \cite{ORE,owdetr,droupoutsampling} and report the performance in Table.\ref{table6}. 
CAT achieves promising performance in comparison to the existing methods.

\section{Relation to Prior Works}
The issue of standard object detection \cite{detection1,detection2,detection3,detection4,detection5,detection6,detection7,fasterrcnn,ddetr,detr,zou2019object} has been raised for several years, numberous works have investigated this problem and push the field to certain heights. Whereas the strong assumption that the label space of object categories to be encountered during the life-span of the model is the same as during its training results that these methods cannot meet real-world needs. The success of \cite{fasterrcnn,OLN,location0,location1,location2,location3, chen2019relation} demonstrates the feasibility of foreground localization based on the position and appearance of objects. Existing works \cite{ORE,owdetr,OCPL,two-branch,mvit,UC-OWOD, chen2022weakly} attempt to leverage the framework of standard object detection models for open-world object detection. In this paper, we propose a novel transformer \cite{transformer} based framework. CAT decouples the localization and identification process and connects them in a cascade approach. In CAT, the foreground localization process is not limited by the category of known objects, whereas the process of foreground identification can use information from the localization process. Along with self-adaptive pseudo-labelling, CAT can gain information beyond the data annotation and maintain a stable learning process according to self-regulation.
\begin{table}[htbp]
\centering
\caption{\textbf{Performance comparison on open-set object detection task}. Our CAT achieves significant performance in comparison to existing works.}
\resizebox{0.7\linewidth}{!}{
\begin{tabular}{l|cc}
\toprule
Evaluated on $\rightarrow$ & VOC &  WR1 \\ \midrule
\multicolumn{1}{l|}{Standard Faster R-CNN\cite{iod}}        & 81.8  & 77.1  \\
\multicolumn{1}{l|}{Standard RetinaNet} & 79.2  & 73.8 \\
\multicolumn{1}{l|}{Dropout Sampling\cite{droupoutsampling}} & 78.1  & 71.1 \\
\multicolumn{1}{l|}{ORE\cite{ORE}}         &81.3  & 78.2 \\
\multicolumn{1}{l|}{OW-DETR\cite{owdetr}}     & 82.1  & 78.6 \\ \midrule
\multicolumn{1}{l|}{\textbf{Ours: CAT}}   &\textbf{83.2}&\textbf{79.5} \\ \bottomrule
\end{tabular}}
\label{table6}
\vspace{-0.4cm}
\end{table}
\section{Conclusions}
We analyze the drawbacks of the parallel decoding structure for open-world object detection. Motivated by the subconscious reactions of humans when facing new scenes, we propose a novel localization and identification cascade detection transformer (CAT), which decouples the localization and identification process via the cascade decoding manner. The cascade decoding manner alleviates the influence of detecting unknown objects on the detection of known objects. With the self-adaptive pseudo-labelling mechanism, CAT gains knowledge beyond the data annotations, generates pseudo labels with robustness and maintains a stable training process via self-adjustment. The extensive experiments on two popular benchmarks, $i.e.$, PASCAL VOC and MS COCO demonstrate that CAT's performance is better than the existing methods. 

\section*{Acknowledgment}
This work is supported by National Nature Science Foundation of China (grant No.61871106), and the Open Project Program Foundation of the Key Laboratory of Opto-Electronics Information Processing, Chinese Academy of Sciences (OEIP-O-202002).

{\small
\bibliographystyle{ieee_fullname}
\bibliography{egbib}
}

\end{document}